\documentclass[runningheads]{llncs}
\usepackage[T1]{fontenc}
\usepackage{graphicx}
\usepackage{booktabs}
\usepackage{multirow}
\usepackage[hidelinks]{hyperref}
\usepackage{xurl}
\usepackage{microtype}

\begin{document}
\title{OUIDecay: Adaptive Layer-wise Weight Decay for CNNs Using Online Activation Patterns}
\titlerunning{OUIDecay: Adaptive WD for CNNs using Activation Patterns}
\author{
Alberto Fernández-Hernández\inst{1} \and
Jose I. Mestre\inst{1} \and
Cristian Pérez-Corral\inst{1} \and
Manuel F. Dolz\inst{2} \and
Jose Duato\inst{3} \and
Enrique S. Quintana-Ortí\inst{1}
}

\authorrunning{A. Fernández-Hernández et al.}

\institute{
Universitat Politècnica de València, Valencia, Spain
\email{\{a.fernandez, jmiravet, cpercor, quintana\}@upv.es}
\and
Universitat Jaume I, Castelló de la Plana, Spain
\email{dolzm@uji.es}
\and
Openchip \& Software Technologies S.L., Spain
\email{jose.duato@openchip.com}
}
\maketitle              % typeset the header of the contribution
\begin{abstract}
Weight decay remains one of the most widely used regularization mechanisms for training convolutional neural networks, yet it is still commonly applied as a fixed coefficient shared by all layers throughout training. This uniform treatment ignores that different layers may follow different structural dynamics and therefore may require different regularization strengths. In this work, we propose OUIDecay, an adaptive layer-wise and time-dependent weight decay scheduler for CNNs driven by the Overfitting-Underfitting Indicator (OUI), an activation-based metric previously shown to provide early information about regularization quality. OUIDecay uses a lightweight batch-based formulation of OUI to monitor the structural behavior of each layer online and periodically rescales its weight decay relative to the other layers in the network. Unlike gradient-based adaptive decay methods, our approach relies on functional information extracted from activation patterns and does not require validation data. Experiments on EfficientNet-B0 with Stanford Cars, ResNet50 with Food101, DenseNet121 with CIFAR100, and MobileNetV2 with CIFAR10 show that OUIDecay achieves the best mean best-validation-loss in 7 out of 8 evaluated settings. These results indicate that activation-driven weight decay adaptation is a practical and effective alternative to fixed decay and gradient-based adaptive decay, while keeping the method lightweight and suitable for online use.

\end{abstract}

\section{Introduction}

Weight decay remains one of the standard tools for regularizing deep neural networks~\cite{krogh_simple_1991}.  Despite its importance, it is still commonly applied in a uniform way: a single coefficient is chosen before training and then shared by all layers for the entire run. This practice is simple and often effective~\cite{dangelo_why_2024}, but it also ignores a basic fact about modern convolutional networks: different layers do not necessarily evolve in the same way, nor do they play the same role throughout optimization. As a result, a uniform decay strength may be unnecessarily restrictive for some parts of the model and too weak for others~\cite{loshchilov_decoupled_2019}.

This limitation has motivated the development of adaptive weight decay strategies. Among them, AdaDecay~\cite{nakamura_adaptive_2019} is especially relevant to our setting. Instead of relying on a fixed decay coefficient, it modulates the regularization strength using normalized gradient information and a sigmoid-based mapping. This is a meaningful departure from static decay, but its control signal still comes from the gradient domain. More broadly, most adaptive approaches adjust regularization through quantities derived from gradients, weights, or optimization geometry, rather than from the internal functional behavior of the network itself.

A different viewpoint is offered by the \emph{Overfitting-Underfitting Indicator} (OUI)~\cite{fernandez-hernandez_oui_2025}, introduced as a label-free metric derived from activation patterns. OUI was proposed to characterize the structural behavior of a network during training, and previous results showed that it can identify, early in training, whether the current weight decay regime is likely to be too weak, too strong, or close to a good regularization balance. This makes OUI attractive for a practical reason. If the metric already carries information about whether the model is drifting toward underfitting or overfitting, then it may serve not only as a diagnostic tool, but also as a control signal for adapting weight decay online.

This observation motivates the present work. We ask a simple question: 

\begin{center}
    \emph{Can OUI be used directly to drive weight decay during CNN training?} 
\end{center}

To answer it, we build on a batch-based formulation of OUI introduced in later work~\cite{fernandez-hernandez_when_2026}, which replaces pairwise comparisons with a compact population-level statistic that is more stable and substantially cheaper to compute online. This makes it possible to monitor the structural state of each layer during training without introducing a heavy computational burden. Our goal is deliberately practical. We do not seek a general theory of weight decay, nor do we assume the existence of a universal target value that every layer should satisfy. Instead, we use OUI as an online functional signal to decide how weight decay should be redistributed across layers over time.

Based on this idea, we propose \emph{OUIDecay}, an adaptive layer-wise and time-dependent weight decay scheduler for CNNs. The method uses layer-level OUI measurements to adjust the base decay coefficient throughout training. Our method is closer in spirit to layer-wise scheduling strategies than to scalar hyperparameter tuning, but it differs from recent layer-wise approaches such as AlphaDecay~\cite{he_alphadecay_2025} in a key aspect: the driving signal is not spectral or weight-based, but activation-based and directly tied to the network's functional behavior.

The contributions of this paper are twofold:
\begin{enumerate}
    \item We propose \emph{OUIDecay}, a layer-wise and time-dependent weight decay scheduler for CNNs driven by the batch-based formulation of OUI.
    \item We show empirically that, in the CNN settings studied in this work, OUIDecay generally improves over fixed weight decay and over AdaDecay.
\end{enumerate}

The remainder of this paper is structured as follows. Section~2 reviews related work on weight decay and adaptive regularization. Section~3 introduces the proposed method and its underlying formulation. Section~4 describes the experimental setup, including the configurations and evaluation protocol. Section~5 presents the main results, followed by an ablation and analysis in Section~6. Finally, Section~7 concludes the paper and discusses possible directions for future work.

\paragraph{Code availability.}
The implementation of OUIDecay, together with the code required to reproduce the experiments reported in this paper, is publicly available at \url{https://github.com/AlbertoFdezHdez/OUIDecay}.

\section{Related Work}

Weight decay is most commonly used as a fixed regularization coefficient shared by all trainable parameters throughout optimization. This remains the default choice in many practical settings, including decoupled formulations such as AdamW~\cite{loshchilov_decoupled_2019}, where the regularization effect is separated from the gradient update but still controlled by a single scalar hyperparameter. While this simple design is often effective, it does not account for the fact that different layers may follow different training dynamics and may therefore benefit from different regularization strengths.

This limitation has motivated a line of work on \emph{adaptive} weight decay. In the context of our paper, the most relevant reference is AdaDecay~\cite{nakamura_adaptive_2019}, which adjusts the regularization strength from gradient information. More precisely, AdaDecay normalizes gradient magnitudes within each layer and maps them through a sigmoid function to obtain parameter-wise decay factors. The method is directly related to our setting because it departs from fixed weight decay while remaining lightweight enough for standard training pipelines. At the same time, it also highlights the main conceptual contrast with our approach: AdaDecay is driven by gradients, whereas OUIDecay is driven by activations. This distinction is central to our work, since we are interested in adapting weight decay from a functional signal that reflects how the network is structurally using its representations during training, rather than from an optimization signal alone.

Our approach builds on previous work on the \emph{Overfitting-Underfitting Indicator} (OUI)~\cite{fernandez-hernandez_oui_2025}, but addresses a different problem. OUI was originally introduced as a label-free diagnostic metric based on binary activation patterns, with the aim of monitoring the structural behavior of deep networks during training. In that setting, OUI was used to analyze fixed weight decay regimes and to provide early evidence about whether a chosen regularization strength was likely to be too weak, too strong, or better balanced. However, the method did not modify the training process itself, nor did it provide a mechanism for assigning different regularization strengths to different layers over time. This distinction is central to our work: the previous results suggested that activation patterns contain useful information about regularization quality, but they left open how such information could be converted into an online control rule. In particular, they did not address how to compute and use the signal efficiently during training, how to aggregate it at the layer level, how to translate relative OUI values into layer-wise decay coefficients, or how often those coefficients should be updated. OUIDecay addresses these missing pieces by turning OUI from a diagnostic indicator into a lightweight online scheduler for adaptive, layer-wise weight decay.

A second OUI-based contribution, developed in the context of reinforcement learning, is also important for the present paper, although for a more specific reason~\cite{fernandez-hernandez_when_2026}. That work introduced a batch-based formulation of OUI that replaces pairwise comparisons with a population-level statistic computed over a probe batch. This reformulation makes OUI more stable and substantially cheaper to evaluate, which is precisely what is needed for online scheduling during training. In addition, the reinforcement learning study showed that early structural signals can discriminate between training regimes well before the end of optimization. We do not use that work as evidence about weight decay in CNNs, because it is not. What we borrow from it is the formulation of the metric and the broader idea that internal structural signals can be informative early enough to guide training decisions.

Finally, our design is also loosely inspired by recent layer-wise scheduling strategies such as AlphaDecay~\cite{he_alphadecay_2025}. AlphaDecay proposes assigning different decay strengths to different layers, using spectral information derived from weight matrices to balance layer-wise behavior in LLM training. The relevance of this work to our paper is limited but clear. It reinforces the idea that a single global decay coefficient may be suboptimal when different layers behave differently, and it offers a useful scheduling perspective based on relative allocation across layers. However, the signal, the model family, and the motivation are all different. AlphaDecay is spectral and weight-based, and it is designed for LLMs. OUIDecay is activation-based, function-oriented, and designed for CNN training. For this reason, AlphaDecay should be understood here as a source of design inspiration rather than as a directly comparable antecedent.

Taken together, these works define the gap addressed in this paper. Fixed weight decay ignores layer heterogeneity. AdaDecay adapts decay from gradients. Previous OUI work showed that activation-based structural signals can reveal whether a weight decay regime is appropriate, and the later batch-based formulation made such signals practical for online monitoring. What remained missing was a method that uses this kind of activation-based functional signal to adapt weight decay online, layer by layer, during CNN training. OUIDecay is designed to fill precisely that gap.

\section{Method}

Our method builds on OUI~\cite{fernandez-hernandez_oui_2025}, a scalar metric in $[0,1]$ designed to quantify the structural variability used by a network when processing data. OUI is label-free, as it depends only on activation patterns, not on targets, losses, or gradients. In the context of this paper, this is precisely its appeal. Rather than inferring whether regularization should be strengthened or relaxed from optimization quantities alone, we use a signal that reflects how each layer is functionally partitioning the input batch.

We adopt the batch-based formulation of OUI introduced in~\cite{fernandez-hernandez_when_2026}, which is more suitable for online monitoring than the original pairwise version. Consider a layer $i$ with $d_i$ scalar activation units and a probe batch $\mathcal{B}_t=\{x_1,\dots,x_B\}$ at training step $t$. Let $a^{(i)}_j(x_b;\theta_t)$ denote the preactivation of unit $j$ in layer $i$ for sample $x_b$, under parameters $\theta_t$. We define the binary activation mask
\begin{equation}
m^{(i)}_{b,j}(t)=\mathbf{1}\!\left\{a^{(i)}_j(x_b;\theta_t)>0\right\},
\end{equation}
for $b\in\{1,\dots,B\}$ and $j\in\{1,\dots,d_i\}$. For each unit $j$, we count how many samples in the batch activate it,
\begin{equation}
s^{(i)}_j(t)=\sum_{b=1}^{B} m^{(i)}_{b,j}(t),
\end{equation}
and then define its \emph{minority count} as
\begin{equation}
u^{(i)}_j(t)=\min\!\left(s^{(i)}_j(t),\, B-s^{(i)}_j(t)\right).
\end{equation}
The batch-based OUI of layer $i$ at step $t$ is finally given by
\begin{equation}
\label{eq:ouibatch}
\mathrm{OUI}_i(t)=
\frac{1}{d_i}
\sum_{j=1}^{d_i}
\frac{u^{(i)}_j(t)}{\lfloor B/2 \rfloor}.
\end{equation}

This definition has a simple operational meaning. $\mathrm{OUI}_i(t)$ is high when many units in layer $i$ split the batch in a roughly balanced way, and it decreases when units become structurally biased, that is, when they are almost always active or almost always inactive. Compared with the pairwise formulation used in the original OUI work~\cite{fernandez-hernandez_oui_2025}, the batch-based version avoids the combinatorial cost of comparing sample pairs and is therefore much better suited to lightweight online use~\cite{fernandez-hernandez_when_2026}.

\emph{OUIDecay} uses these layer-level OUI values to adapt weight decay during training. Let $\lambda_{base}$ denote the base weight decay chosen for the run, and let $\tilde{t}$ be the update interval of the scheduler. Every $\tilde{t}$ optimization steps, we compute $\mathrm{OUI}_i(t)$ for each layer $i\in\{1,\dots,M\}$ under consideration and then extract the layer-wise extrema
\begin{equation}
\mathrm{OUI}_{\min}(t)=\min_i \mathrm{OUI}_i(t),
\qquad
\mathrm{OUI}_{\max}(t)=\max_i \mathrm{OUI}_i(t).
\end{equation}
These values are used to assign a layer-wise weight decay coefficient through a linear rescaling:
\begin{equation}
\label{eq:ouidecay}
\lambda_i(t)=
\lambda_{base} \left[
s_1 + (s_2-s_1) \,
\frac{\mathrm{OUI}_i(t)-\mathrm{OUI}_{\min}(t)}
{\mathrm{OUI}_{\max}(t)-\mathrm{OUI}_{\min}(t)+\varepsilon}
\right],
\end{equation}
where $(s_1,s_2)$ defines the scaling range applied to the base weight decay $\lambda_{base}$, and $\varepsilon>0$ is a small constant for numerical stability. Therefore, the weight decay assigned to module $i$ is constrained to lie between $s_1\lambda_{base}$ and $s_2\lambda_{base}$. The exact value within this interval is determined by the relative position of $\mathrm{OUI}_i(t)$ with respect to the other monitored modules at the same update step: modules with OUI values closer to $\mathrm{OUI}_{\min}(t)$ receive coefficients closer to $s_1\lambda_{base}$, whereas modules with OUI values closer to $\mathrm{OUI}_{\max}(t)$ receive coefficients closer to $s_2\lambda_{base}$. Between scheduler updates, the assigned coefficients are kept unchanged.

The purpose of the scheduler can be stated plainly: \emph{to balance OUI across layers during training through a relative adaptation of weight decay}. This point matters because OUIDecay does not seek an absolute target value of OUI. We deliberately avoid prescribing that every layer should approach a single ``ideal'' OUI level. Such a design would hard-code a target that may depend on the architecture, the dataset, the optimizer, or even the training stage. Instead, our method only uses the relative ordering and spread of OUI values across layers at a given time. In other words, OUIDecay reacts to \emph{imbalance} rather than to deviation from a global target.

This relative design also explains why the scheduler is not updated at every iteration. Indeed, a per-step update would make the assigned decay coefficients more sensitive to short-term fluctuations in the batch statistics. Updating every $\tilde{t}$ steps provides a more stable signal, keeps the method easy to integrate into standard pipelines, and preserves the lightweight character that motivates the whole approach.

The intuition behind OUIDecay is that, if one layer exhibits a structural dynamics that differs markedly from the rest of the network, then a uniform weight decay value may not be equally appropriate for all layers at that moment. OUIDecay uses this observable difference in activation behavior to redistribute regularization across layers over time. In this sense, the method is not trying to infer hidden properties of the loss landscape. It simply uses an internal functional signal to decide where stronger or weaker regularization may be more appropriate during training.

For clarity, the scheduling procedure is summarized below.

\begin{center}
\fbox{
\begin{minipage}{0.94\linewidth}
\small
\textbf{Algorithm 1: OUIDecay}

\medskip
\textbf{Input:} base weight decay $\lambda_{base}$, scheduler interval $\tilde{t}$, scaling range $(s_1,s_2)$, stability constant $\varepsilon$.

\medskip
For each training step $t$:
\begin{enumerate}
    \item Perform the usual forward and backward pass.
    \item If $t \bmod \tilde{t} = 0$:
    \begin{enumerate}
        \item Compute $\mathrm{OUI}_i(t)$ for every monitored layer $i$ using Equation (\ref{eq:ouibatch}).
        \item Compute $\mathrm{OUI}_{\min}(t)$ and $\mathrm{OUI}_{\max}(t)$ across layers.
        \item Assign layer-wise decay coefficients $\lambda_i(t)$ using Equation (\ref{eq:ouidecay}).
    \end{enumerate}
    \item Otherwise, keep the most recent $\lambda_i$ values unchanged.
    \item Update the network parameters using the optimizer with the current layer-wise weight decay coefficients.
\end{enumerate}
\end{minipage}
}
\end{center}

\section{Experimental Evaluation}
This section evaluates whether \emph{OUIDecay} improves the final training outcome over fixed weight decay and AdaDecay. We first describe the experimental protocol used for the comparison, then report the quantitative results across the selected CNN settings, and finally analyze the main design choices of the method through targeted ablations and runtime measurements.

\subsection{Experimental Protocol}

Our experiments are designed to answer whether \emph{OUIDecay} improves the final training outcome relative to \emph{fixed weight decay} and \emph{AdaDecay}~\cite{nakamura_adaptive_2019}. The main body of the paper focuses on four representative CNN settings:
\emph{EfficientNet-B0} on \emph{Stanford Cars}~\cite{tan_efficientnet_2019,krause_3d_2013},
\emph{ResNet50} on \emph{Food101}~\cite{he_deep_2016,bossard_food-101_2014},
\emph{DenseNet121} on \emph{CIFAR100}~\cite{huang_densely_2017,krizhevsky_learning_2009},
and \emph{MobileNetV2} on \emph{CIFAR10}~\cite{sandler_mobilenetv2_2018,krizhevsky_learning_2009}. These configurations were selected from the final CNN sweep because together they cover different backbone families, image scales, and regularization regimes while keeping the study compact and easy to interpret.

Our primary metric is the \emph{best validation loss}. For each configuration, we report the mean and standard deviation of the best validation loss over seeds 1, 2, and 3. We treat this quantity as the main evaluation criterion because it is the most directly aligned with the role of weight decay as a regularizer, and because it proved more stable and informative than accuracy for the comparisons that matter in this paper.

In all configurations, we evaluate two base weight decay values separated by a multiplicative factor of five. This design choice is intended to make the comparison more robust and to avoid favoring adaptive methods through an arbitrary initial scaling. Since OUIDecay redistributes the base coefficient within a range that can span multiple times its original value, evaluating a single baseline could lead to ambiguous conclusions if the chosen weight decay were suboptimal. By explicitly including both the nominal value and a $5\times$ scaled version, we ensure that the comparison accounts for different regularization regimes. This allows us to verify that the observed improvements are not simply due to an implicit rescaling effect, but persist across distinct and reasonably spaced weight decay settings.

Across all settings, learning rate follows a warmup-plus-cosine schedule, while the optimizer, base weight decay, batch size, training length, and clipping behavior remain specific to each model-dataset pair. Table~\ref{tab:setup} summarizes the main training configurations used in the paper. 

\begin{table*}[ht]
\centering
\caption{Main experimental configurations used in the paper. For each model-dataset pair, we report the optimizer, the base weight decay values explored in the final comparison (Base WDs), the training length (Epochs), and the batch size (BS).}
\label{tab:setup}
\renewcommand{\arraystretch}{1.15}
\setlength{\tabcolsep}{6pt}
\begin{tabular*}{\textwidth}{@{\extracolsep{\fill}} l l c c c c @{}}
\toprule
\textbf{Model} & \textbf{Dataset} & \textbf{Optimizer} & \textbf{Base WDs} & \textbf{Epochs} & \textbf{BS} \\
\midrule
EfficientNet-B0 & Stanford Cars & Adam  & $10^{-5},\,5\times10^{-5}$ & 100 & 64 \\
ResNet50        & Food101       & AdamW & $10^{-2},\,5\times10^{-2}$ & 50  & 128 \\
DenseNet121     & CIFAR100      & AdamW & $5\times10^{-2},\,10^{-1}$ & 100 & 256 \\
MobileNetV2     & CIFAR10       & Adam  & $10^{-4},\,5\times10^{-4}$ & 100 & 256 \\
\bottomrule
\end{tabular*}
\end{table*}

The remaining elements of the training recipe are also held fixed within each configuration. First, the learning rate follows the same warmup-plus-cosine schedule but the base and minimum values are configuration-specific: EfficientNet-B0 uses $8\times10^{-4}\rightarrow10^{-5}$, ResNet50 uses $3\times10^{-4}\rightarrow3\times10^{-5}$, and both DenseNet121 and MobileNetV2 use $5\times10^{-4}\rightarrow5\times10^{-6}$. On the other hand, EfficientNet-B0 on Stanford Cars uses the strongest augmentation pipeline, with random resized crops, horizontal flips, color jitter, RandAugment, and random erasing, while validation uses resize followed by center crop. ResNet50 on Food101 uses random resized crops, horizontal flips, color jitter, and standard normalization, again with resize and center crop at validation time. DenseNet121 on CIFAR100 and MobileNetV2 on CIFAR10 share the same CIFAR-style augmentation strategy, namely random crop with padding, horizontal flip, RandAugment, and random erasing, whereas validation uses only normalization. Explicit gradient clipping is only applied in the EfficientNet-B0 setting, with clipping norm set to 1.0; the other configurations use no clipping. 

For \emph{OUIDecay}, layers are defined at the level used by the corresponding backbone decomposition in the implementation, so that each monitored unit corresponds to a meaningful trainable block rather than to an excessively fine parameter partition. Unless stated otherwise, we use $s_1=0.67$, $s_2=5$, and an update interval of $\tilde{t}=500$ iterations. The scheduler is therefore not refreshed at every step, but periodically, which keeps the method lightweight and stable in practice. In the ablation study, we vary both the update interval $\tilde{t}$ and the scaling range $(s_1,s_2)$ in order to quantify the trade-off between responsiveness, stability, and computational cost.

\subsection{Comparison on Best Validation Loss}

Table~\ref{tab:main_results} reports the main quantitative comparison of the paper. The result is clear at a glance: \emph{OUIDecay} achieves the best mean best validation loss in 7 of the 8 reported settings.

\begin{table*}[ht]
\centering
\caption{Main comparison in terms of best validation loss (mean $\pm$ standard deviation over seeds 1, 2, and 3). Lower is better. The best mean in each row is shown in bold.}
\label{tab:main_results}
\renewcommand{\arraystretch}{1.14}
\setlength{\tabcolsep}{5pt}
\resizebox{\textwidth}{!}{%
\begin{tabular}{l l c c c c}
\toprule
\textbf{Network} & \textbf{Dataset} & \textbf{WD$_0$} & \textbf{Fixed WD} & \textbf{AdaDecay} & \textbf{OUIDecay} \\
\midrule
EfficientNet-B0 & Stanford Cars & $10^{-5}$ &
$1.27 \pm 0.03$ &
$1.29 \pm 0.02$ &
$\mathbf{1.17 \pm 0.02}$ \\

EfficientNet-B0 & Stanford Cars & $5\times10^{-5}$ &
$1.20 \pm 0.03$ &
$1.21 \pm 0.03$ &
$\mathbf{1.06 \pm 0.02}$ \\

ResNet50 & Food101 & $10^{-2}$ &
$\mathbf{1.03 \pm 0.01}$ &
$1.12 \pm 0.03$ &
$\mathbf{1.03 \pm 0.00}$ \\

ResNet50 & Food101 & $5\times10^{-2}$ &
$1.02 \pm 0.01$ &
$1.58 \pm 0.16$ &
$\mathbf{1.00 \pm 0.01}$ \\

DenseNet121 & CIFAR100 & $5\times10^{-2}$ &
$1.37 \pm 0.01$ &
$\mathbf{1.17 \pm 0.01}$ &
$1.33 \pm 0.01$ \\

DenseNet121 & CIFAR100 & $10^{-1}$ &
$1.35 \pm 0.01$ &
$1.29 \pm 0.00$ &
$\mathbf{1.28 \pm 0.00}$ \\

MobileNetV2 & CIFAR10 & $10^{-4}$ &
$0.39 \pm 0.00$ &
$0.40 \pm 0.00$ &
$\mathbf{0.36 \pm 0.00}$ \\

MobileNetV2 & CIFAR10 & $5\times10^{-4}$ &
$0.35 \pm 0.00$ &
$0.36 \pm 0.00$ &
$\mathbf{0.33 \pm 0.00}$ \\
\bottomrule
\end{tabular}%
}
\end{table*}

The most consistent gains appear in the EfficientNet-B0 experiments on Stanford Cars. In both base weight decay settings, OUIDecay improves clearly over the two baselines, and the gap is not marginal. This matters because the trend is repeated across both rows, which makes the result less likely to be an isolated effect of a single regularization value. In this configuration, the advantage of using activation-driven scheduling is both visible and stable.

The ResNet50 experiments on Food101 show a more nuanced picture, and that nuance is worth preserving. For the smaller base weight decay, OUIDecay and fixed weight decay are nearly indistinguishable in practice: the difference in mean best validation loss is negligible. For the larger base weight decay, however, OUIDecay improves clearly over both alternatives. This is also one of the settings where AdaDecay appears more sensitive, with a substantially larger standard deviation and a markedly worse mean. 

The DenseNet121 results on CIFAR100 are mixed, but still informative. For the lower of the two base weight decay values, AdaDecay yields the best result, and this is the only row in the table where OUIDecay does not outperform both baselines. For the larger base weight decay, OUIDecay recovers the lead.

The MobileNetV2 experiments on CIFAR10 return to the clearest pattern in the table. OUIDecay is the best method for both base weight decay values, improving over fixed decay and over AdaDecay in both cases. 

A broader conceptual point emerges from these results. AdaDecay is already a meaningful adaptive baseline, since it adjusts regularization online from gradient information~\cite{nakamura_adaptive_2019}. The fact that OUIDecay surpasses it in most of the studied settings supports the hypothesis that, at least in these experiments, a \emph{functional} signal based on activations can be more useful for weight decay adaptation than a signal derived from gradients alone. We emphasize the scope of this claim deliberately: it applies to the CNN settings reported here, not to every possible training regime.

Overall, the main result of the paper is that OUIDecay improves the final quality of training in a consistent and practically relevant way across several CNN backbones and datasets, while preserving the lightweight character that motivated the method from the start.

\subsection{Comparison on Best Validation Loss}

This section examines whether the behavior of \emph{OUIDecay} follows from its design choices or is merely an incidental consequence of perturbing the weight decay coefficient during training. We focus on three concrete aspects: the scheduler update interval, the scaling range used to redistribute the base weight decay, and the computational cost of computing the OUI signal. 

\paragraph{Ablation on the update interval.}
The first design choice concerns the scheduler update interval $\tilde{t}$, which determines how often the layer-wise OUI signal is refreshed. Intuitively, this parameter plays two roles at once. On the one hand, smaller values allow the scheduler to react more quickly to changes in the structural state of the model. On the other hand, updating too often may introduce unnecessary short-term variability, effectively over-correcting the regularization strength before the network has time to absorb the previous adjustment. In that sense, $\tilde{t}$ acts not only as a refresh rate, but also as a stabilizer.

\begin{table}[ht]
\centering
\caption{Ablation on the scheduler update interval $\tilde{t}$ for EfficientNet-B0 on Stanford Cars with Adam, base weight decay $5\times 10^{-5}$, seed 1, and OUIDecay. Lower best validation loss is better.}
\label{tab:ablation_tgap}
\renewcommand{\arraystretch}{1.14}
\setlength{\tabcolsep}{4pt}

\begin{tabular*}{\columnwidth}{@{\extracolsep{\fill}} c c c c @{\hspace{3em}} c c c c @{}}
\toprule
\textbf{$t$} & \textbf{$\tilde{t}$} & \textbf{Best val. loss $\downarrow$} & \textbf{Time} &
\textbf{$t$} & \textbf{$\tilde{t}$} & \textbf{Best val. loss $\downarrow$} & \textbf{Time} \\
\midrule
$2^0$  & $1$   & $1.1029$           & $2472.94\,\mathrm{s}$ &
$2^7$  & $128$ & $\mathbf{1.0004}$ & $2372.60\,\mathrm{s}$ \\

$2^2$  & $4$   & $1.0758$           & $2451.38\,\mathrm{s}$ &
$2^8$  & $256$ & $\textbf{1.0085}$          & $2372.76\,\mathrm{s}$ \\

$2^4$  & $16$  & $1.0448$           & $2444.97\,\mathrm{s}$ &
$2^9$  & $512$ & $\textbf{1.0026}$          & $2354.47\,\mathrm{s}$ \\

$2^6$  & $64$  & $1.0342$           & $2450.66\,\mathrm{s}$ &
$2^{10}$ & $1024$ & $1.0225$       & $2541.19\,\mathrm{s}$ \\
\bottomrule
\end{tabular*}
\end{table}

Table~\ref{tab:ablation_tgap} reports a sweep for $\tilde{t} \in \{1, 4, 16, 64, 128, 256, 512, 1024\}$ on EfficientNet-B0 with Stanford Cars, using Adam, base weight decay $5\times 10^{-5}$, seed 1, and OUIDecay. The trend is clear. Very frequent updates are not beneficial, and the best results appear in an intermediate regime. The strongest values in this sweep are obtained at $\tilde{t}$ between $128$ and $512$, which motivates the selection of the value $\tilde{t} = 500$ which is the default choice in the rest of the experiments: it stays near the best region of the sweep while being slightly more conservative, which helps avoid excessive oscillation in the scheduler. Additionally, the total runtime does not show the strong inverse dependence on $\tilde{t}$ that one might expect if OUI computation were a major bottleneck. This already suggests that the cost of the scheduler is small relative to the cost of the training iteration itself.

\paragraph{Computational overhead of OUI updates.}
We now quantify that cost more directly. Table~\ref{tab:overhead} reports the time required for a full training iteration and the extra time associated with one OUI and weight-decay update in each of the four main experimental settings. In all cases, the additional cost is below $0.2\%$ of the full iteration time. This is negligible in practice, and once updates are performed only every $\tilde{t}$ iterations, the amortized impact becomes even smaller.

\begin{table}[ht]
\centering
\caption{Time cost of one OUI and weight-decay update relative to the full training iteration.}
\label{tab:overhead}
\renewcommand{\arraystretch}{1.14}
\setlength{\tabcolsep}{4.5pt}
\begin{tabular*}{\columnwidth}{@{\extracolsep{\fill}} l l c c c @{}}
\toprule
\textbf{Model} & \textbf{Dataset} & \textbf{Full iter.} & \textbf{OUI/WD update} & \textbf{\% of iter.} \\
\midrule
EfficientNet-B0 & Stanford Cars & $689.43\,\mathrm{ms}$ & $0.2024\,\mathrm{ms}$ & $0.029\%$ \\
ResNet50        & Food101       & $511.82\,\mathrm{ms}$ & $0.8581\,\mathrm{ms}$ & $0.168\%$ \\
DenseNet121     & CIFAR100      & $168.54\,\mathrm{ms}$ & $0.2572\,\mathrm{ms}$ & $0.153\%$ \\
MobileNetV2     & CIFAR10       & $95.78\,\mathrm{ms}$  & $0.1356\,\mathrm{ms}$ & $0.142\%$ \\
\bottomrule
\end{tabular*}
\end{table}

\paragraph{Ablation on the scaling range.}
The remaining design choice is the scaling interval $(s_1,s_2)$ used to redistribute the base weight decay across layers. This parameter controls how much flexibility the scheduler has when separating layers with different OUI values. If the interval is too narrow, OUIDecay becomes too similar to fixed weight decay and loses most of its expressive power. If the interval is too aggressive, the assigned coefficients may become unnecessarily extreme. Table~\ref{tab:ablation_scaling} reports this ablation on ResNet50 with Food101 and AdamW.

\begin{table}[ht]
\centering
\caption{Ablation on the scaling range $(s_1,s_2)$ for ResNet50 on Food101 with AdamW. Lower validation loss is better.}
\label{tab:ablation_scaling}
\renewcommand{\arraystretch}{1.14}
\setlength{\tabcolsep}{4pt}

\begin{tabular*}{\columnwidth}{@{\extracolsep{\fill}} c c @{\hspace{3em}} c c @{}}
\toprule
\textbf{$(s_1,s_2)$} & \textbf{Val. loss $\downarrow$} &
\textbf{$(s_1,s_2)$} & \textbf{Val. loss $\downarrow$} \\
\midrule
$(0.67,\,5.0)$ & $\mathbf{0.9951 \pm 0.0146}$ &
$(0.33,\,3.0)$ & $1.0135 \pm 0.0112$ \\

$(0.67,\,3.0)$ & $1.0083 \pm 0.0165$ &
$(0.33,\,5.0)$ & $0.9962 \pm 0.0213$ \\

\bottomrule
\end{tabular*}
\end{table}

The differences are not dramatic, which is itself a useful result: the method is not hypersensitive to the exact choice of scaling interval. At the same time, the best result is obtained with $(s_1,s_2)=(0.67,5.0)$, which we therefore adopt as the default configuration.

\section{Conclusion}

We have presented \emph{OUIDecay}, a layer-wise and time-dependent weight decay scheduler for CNNs driven by OUI, an activation-based signal that can be monitored online during training. The method is built around a simple idea: if different layers exhibit different structural dynamics, then a uniform regularization strength may be unnecessarily rigid, and an internal functional signal can be used to redistribute weight decay more appropriately over time.

This perspective matters for several reasons. First, OUIDecay relies on activations rather than on gradients or spectral measurements, which gives it a different point of view on regularization from that of existing adaptive schedulers such as AdaDecay~\cite{nakamura_adaptive_2019} or layer-wise spectral methods such as AlphaDecay~\cite{he_alphadecay_2025}. Second, the signal used by the method is internal and label-free, so it does not require validation data to guide the scheduling process. Third, the scheduler remains lightweight and easy to integrate into standard training pipelines, since it uses the batch-based formulation of OUI and updates the layer-wise coefficients only periodically. Within the CNN settings studied in this work, these design choices translate into practical gains over fixed weight decay and, in most cases, over AdaDecay.

At the same time, the scope of the paper is deliberately bounded. Our empirical validation is restricted to CNNs, and the experimental scale is intentionally focused rather than exhaustive. We do not claim that the same behavior will necessarily transfer to transformers or to other model families, nor do we claim that OUIDecay solves the broader problem of weight decay selection in general. The contribution of the paper is narrower and, we believe, more useful for that reason: it shows that activation-based structural signals can serve as a practical basis for online regularization scheduling in convolutional networks.

There are several natural directions for future work. One is to extend the study to other architectures and training regimes in order to test how far the activation-driven perspective carries beyond CNNs. Another is to explore alternative allocation rules built on top of OUI, including non-linear schedules or strategies that incorporate additional temporal information from the evolution of the metric itself. These extensions are worth investigating, but they should build on the same principle that motivates the present work: keeping the method simple enough to remain useful in practice.

In this sense, OUIDecay should be understood as a practical and effective step toward more flexible regularization in CNN training, rather than as a definitive solution to the weight decay problem. Across the settings considered in this work, it consistently improves training outcomes while remaining lightweight and easy to integrate. Its main contribution is to show that online structural information derived from activations can be turned into a simple yet powerful scheduling mechanism, opening a promising direction for activation-driven optimization strategies.

\begin{credits}
\subsubsection{\ackname} This research was funded by the projects PID2023-146569NB-C21 and PID2023-146569NB-C22 supported by MICIU/AEI/10.13039/501100011033 and ERDF/UE. Alberto Fernández-Hernández was supported by the predoctoral grant PREP2023-001826 supported by MICIU/AEI/10.13039/501100011033 and ESF+. Jose I. Mestre was supported by the predoctoral grant ACIF/2021/281 of the \emph{Generalitat Valenciana}. Cristian Pérez-Corral received support from the \textit{Conselleria de Educación, Cultura, Universidades y Empleo} (reference CIACIF/2024/412) through the European Social Fund Plus 2021–2027 (FSE+) program of the \textit{Comunitat Valenciana}. Manuel F. Dolz was supported by grant {\small CNS2025-165098} funded by {\small MICIU/AEI/10.13039/501100011033} and by the Plan Gen--T grant {\small CIDEXG/2022/013} of the \emph{Generalitat Valenciana}.

\subsubsection{\discintname}
The authors have no competing interests to declare that are
relevant to the content of this article.
\end{credits}

\bibliographystyle{splncs04}
\bibliography{references}

\end{document}